\documentclass{article}
\usepackage{authblk}
\usepackage[utf8]{inputenc}
\usepackage[numbers]{natbib}
\usepackage{url}
\usepackage{hyperref}
\usepackage{graphicx}
\usepackage{xcolor}
\usepackage{enumitem}
\usepackage[a4paper, margin=1in]{geometry} 
\usepackage{makecell}
\usepackage{booktabs}

\title{From Mind to Machine: The Rise of Manus AI as a Fully Autonomous Digital Agent}
\author[1]{Minjie Shen}
\author[2]{Yanshu Li}
\author[1]{Lulu Chen}
\author[3]{Zhichao Fan}
\author[4]{Yanhang Li}
\author[3]{Qikai Yang}
\author[5]{Haochen Yang}
\affil[1]{Department of Electrical and Computer Engineering, Virginia Tech}
\affil[2]{Department of Computer Science, Brown University }
\affil[3]{Department of Computer Science, University of Illinois at Urbana-Champaign}
\affil[4]{Northeastern University}
\affil[5]{Harvard University}

\date{}

\begin{document}
\maketitle

\begin{abstract}
\textbf{Manus AI} is a general-purpose AI agent introduced in early 2025 as a breakthrough in autonomous artificial intelligence. Developed by the Chinese startup Monica.im, Manus is designed to bridge the gap between "mind" and "hand" – it not only thinks and plans like a large language model, but also executes complex tasks end-to-end to deliver tangible results. This paper provides a comprehensive overview of Manus AI, examining its underlying technical architecture, its wide-ranging applications across industries (including healthcare, finance, manufacturing, robotics, gaming, and more), as well as its advantages, limitations, and future prospects. Ultimately, Manus AI is positioned as an early glimpse into the future of AI – one where intelligent agents could revolutionize work and daily life by turning high-level intentions into actionable outcomes, auguring a new paradigm of human-AI collaboration.
\end{abstract}

\section{Introduction}


Recent years have witnessed tremendous breakthroughs in artificial intelligence (AI), from the rise of deep neural networks to large language models that can converse and solve complex problems. Models like OpenAI's GPT-4~\cite{OpenAI2023} have demonstrated unprecedented language understanding, yet such systems typically operate as assistants that respond to queries rather than autonomously acting on tasks. The next evolution in AI is the development of \emph{general-purpose AI agents} that can bridge the gap between decision-making and action. \textbf{Manus AI} is a prominent new example, described as one of the world's first truly autonomous AI agents capable of ``thinking" and executing tasks much like a human assistant~\cite{MalayMail2025}. 

Manus AI, developed by the Chinese startup Monica in 2025, has quickly drawn global attention for its ability to perform a wide array of real-world jobs with minimal human guidance. Unlike traditional chatbots that strictly provide information or suggestions, Manus can plan solutions, invoke tools, and carry out multi-step procedures on its own~\cite{EconomicTimes2025}. For example, rather than just giving travel advice, Manus can autonomously plan an entire trip itinerary, gather relevant information from the web, and present a finalized plan to the user, all without step-by-step prompts~\cite{EconomicTimes2025}. This agent-centric approach represents a significant leap in AI capabilities and has fueled speculation that systems like Manus herald the next stage in AI evolution toward artificial general intelligence (AGI).

In benchmark evaluations for general AI agents, Manus AI has reportedly achieved \emph{state-of-the-art} results. On the GAIA test—a comprehensive benchmark assessing an AI's ability to reason, use tools, and automate real-world tasks—Manus outperformed leading models including OpenAI's GPT-4~\cite{LLMHacker2025}. In fact, early reports suggest Manus exceeded the previous GAIA leaderboard champion's score of 65\%, setting a new performance record~\cite{LLMHacker2025}. Such achievements underscore the importance of Manus AI as a breakthrough system in the competitive landscape of AI.

This paper provides a detailed examination of Manus AI. Section 2 explains how Manus AI works, delving into its model architecture, core algorithms, training process, and unique features. Section 3 explores Manus AI's applications across various industries—ranging from healthcare and finance to robotics and education—illustrating its versatility. In Section 4, we compare Manus AI with other cutting-edge AI technologies (including offerings from OpenAI, Google DeepMind, and Anthropic) to analyze how Manus stands out. Section 5 discusses the strengths of Manus AI as well as its limitations and ongoing challenges. Section 6 considers future prospects for Manus AI and its broader implications for the field. Finally, Section 7 concludes with a summary of findings and reflections on Manus AI's significance in the trajectory of AI development.

\begin{table}[h]
\centering
\begin{tabular}{lcccc}
\toprule 
\textbf{Feature} & \makecell{\textbf{Manus AI} \\  Monica} & \makecell{\textbf{Operator} \\ OpenAI}
 & \makecell{\textbf{Computer Use} \\ Anthropic} & \makecell{\textbf{Mariner} \\ \makecell{Google}} \\
\hline
\makecell{Agent Type} & \makecell{Browser-based \\ (operates in \\ Linux sandboxs)} & Browser-based & API-based & \makecell{Browser-based \\ (Chrome extension)} \\
\hline
\makecell{Autonomous \\ web browsing} & Yes & Yes & Yes* & Yes \\
\hline
\makecell{Form filling \\ and data entry} & Yes & Yes & Yes* & Yes \\
\hline
\makecell{Online shopping \\ and reservations} & Yes & Yes & Yes* & Yes \\
\hline
\makecell{Multi-modal \\ input/output \\ (text, images)} & Yes & Limited & Limited* & Yes \\
\hline
\makecell{Integration with \\ external APIs} & No & No & Yes & N/A \\
\hline
\makecell{Availability} & \makecell{Beta \\ (invite-only)} & Subscribers & \makecell{Beta \\ (API access)} & Research phase \\
\bottomrule  
\end{tabular}
\caption{Feature comparison of Manus AI, OpenAI's Operator, Anthropic's Computer Use, and Google's Mariner. Note: Features marked with * require integration through the API.}
\label{tab:ai_agents_comparison}
\end{table}


\section{How Manus AI Works}
\subsection*{Architecture and Model Design}

\begin{figure}[h]
    \centering
\includegraphics[width=\textwidth]{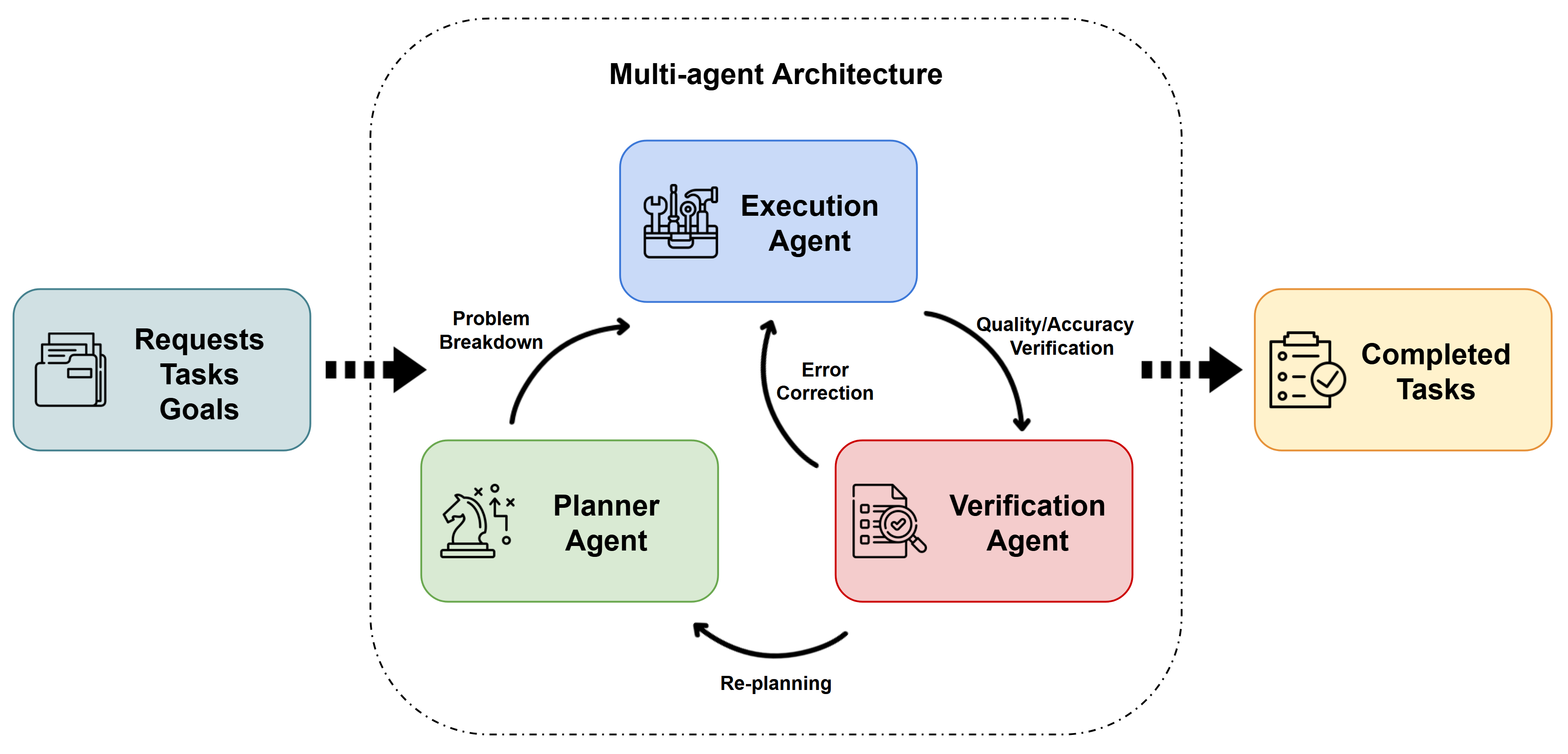}
\caption{Architecture and Model Design}
\end{figure}

Manus AI is built on a sophisticated architecture that combines large-scale machine learning models with an intelligent agent framework. At its core is a transformer-based \emph{large language model} (LLM) that has been trained on vast amounts of textual and multi-modal data. This core model provides the general intelligence, language understanding, and reasoning ability of Manus. However, Manus AI goes beyond a single model by employing a \textbf{multi-agent architecture} that organizes its cognitive processes into specialized modules~\cite{PandaYoo2025}. In particular, Manus consists of at least three coordinated agents working in concert:
\begin{itemize}[leftmargin=*]
    \item \textbf{Planner Agent}: This module functions as the strategist. When a user gives a request or goal, the Planner breaks the problem down into manageable sub-tasks and formulates a step-by-step plan or strategy to achieve the desired outcome.
    \item \textbf{Execution Agent}: This is the action module. The Execution agent takes the Planner's plan and carries it out by invoking the necessary operations or tools. It interacts with external systems (for example, web browsers, databases, code execution environments) to gather information, perform calculations, or execute commands needed for each sub-task.
    \item \textbf{Verification Agent}: Acting as quality control, this module reviews and verifies the outcomes of the Execution agent's actions. It checks results for accuracy and completeness, ensuring that each step meets the requirements before finalizing the output or moving on. The Verification agent can correct errors or trigger re-planning if needed.
\end{itemize}

This multi-agent system runs within a controlled runtime environment (a kind of cloud-based sandbox), essentially creating a “digital workspace” for each task request. By dividing responsibilities among Planner, Execution, and Verification sub-agents~\cite{li2024agent}, Manus AI achieves a level of efficiency and parallelism in task handling. Complex jobs can be tackled by decomposing them and processing components simultaneously, which accelerates completion time compared to a single monolithic model. The architecture is analogous to a small team: one agent plans, another executes, and a third reviews, enabling robust and reliable performance even on complicated, multi-step tasks.

\subsection*{Algorithms and Training Process}
The intelligence of Manus AI's agents is powered by advanced machine learning algorithms. The system leverages deep neural networks for natural language understanding and decision-making, and it has been refined through techniques like \emph{reinforcement learning} to operate effectively in open-ended scenarios~\cite{ChinaOrg2025}. Unlike AI systems that follow fixed rules or only respond to static training data, Manus adapts to unfamiliar situations in real time. During development, the Manus team likely trained the model on a wide range of task demonstrations and used reinforcement learning from human feedback (RLHF)~\cite{ouyang2022training} to align its actions with desired outcomes. This approach allows Manus to dynamically adjust its strategy when encountering new problems, guided by a reward mechanism for successfully completed objectives~\cite{ChinaOrg2025}.

One distinguishing aspect of Manus AI is its \textbf{context-aware decision making}. Rather than executing single-step commands, Manus maintains an internal memory of context and intermediate results as it works through a problem. This means it can take into account the evolving state of a task and user-specific preferences when deciding the next action. The underlying models use sequence-to-sequence predictions to determine the most logical next step, and they update an internal plan as new information is obtained. Manus's algorithms incorporate elements of \emph{human-like reasoning}, attempting to infer what a user ultimately wants and making judgment calls to meet those goals~\cite{ChinaOrg2025}. For example, if a user asks Manus to “analyze sales data and suggest strategies,” Manus will not only compute trends but also decide what types of analyses and visualizations are relevant, and then proceed to generate actionable insights, much as a human analyst might.

To support such complex behavior, Manus AI's training likely involved multi-modal and multitask learning. Reports indicate Manus can handle text, images, and even audio or code as inputs and outputs~\cite{ChinaOrg2025,LLMHacker2025}. This was made possible by training the model on diverse data (e.g. documents, pictures, programming code) and by using a scalable neural network architecture that can fuse information from different modalities. The result is an AI agent capable of interpreting a medical image, reading a scientific article, writing a block of code, and cross-referencing these heterogeneous inputs within a single workflow if a task requires it.

Another key component is Manus AI's \textbf{tool integration capability}. The Execution agent is designed to interface with external applications and APIs. During training, Manus was equipped with the ability to call functions or tools using natural language (a concept similar to “tool use” in other AI agents). For instance, if part of a plan requires getting up-to-date stock prices, Manus knows to invoke a web browsing tool to retrieve the data~\cite{LLMHacker2025}. If the task involves working with structured data, Manus can use a database query tool or a spreadsheet editor. This extensible tool-use framework was likely developed by fine-tuning Manus on examples of how to use various tools and by incorporating APIs for external services. It allows Manus to extend its capabilities beyond what is stored in its neural weights, giving it access to real-time information and specialized functions (like running code or searching the internet) on-the-fly~\cite{LLMHacker2025}.

\subsection*{Unique Features and Capabilities}
Through its architecture and training, Manus AI exhibits several unique features that distinguish it from conventional AI assistants:
\begin{itemize}
    \item \textbf{Autonomous Task Execution}: Manus AI can carry out complex sequences of actions with minimal user intervention. Once given a high-level goal, it will plan, execute, and finalize the task largely on its own. This goes far beyond the typical AI, which would require the user to break down the problem or confirm each step. Manus “excels at various tasks in work and life, getting everything done while you rest,” as its creators put it~\cite{MalayMail2025}. For example, it can generate a detailed report (with visuals and text) from raw data entirely autonomously, or perform all steps of booking a trip after a user simply requests a vacation plan.
    \item \textbf{Multi-Modal Understanding}: Manus AI~\cite{LLMHacker2025} is designed to process and generate multiple types of data, including:
    \begin{itemize}
        \item Text (e.g., generating reports, answering queries) 
        \item Images (e.g., analyzing visual content) 
        \item Code (e.g., automating programming tasks)
    \end{itemize}
    This versatility means Manus can tackle tasks like reading a diagram or X-ray and then writing an explanation of it, or debugging a piece of software based on both the code and error screenshots.
    \item \textbf{Advanced Tool Use}: Manus AI is adept at integrating with external tools and software applications to augment its abilities. It has built-in support for web browsing, so it can fetch up-to-the-minute information from the internet. It can interface with productivity software (for instance, creating or editing spreadsheets and documents) and query databases. This ability to interact with external applications makes Manus AI an ideal tool for businesses looking to automate workflows. Integrating tool use into an AI agent is challenging, and Manus's effective tool usage is a major innovation in bridging AI with practical automation tasks.
    \item \textbf{Continuous Learning and Adaptation}: Manus AI continuously learns from user interactions and optimizes its processes to provide personalized and efficient responses. This ensures that over time, the AI becomes more tailored to the specific needs of the user~\cite{LLMHacker2025}. For example, if a user consistently prefers data presented in a certain format or tone, Manus will adapt to those preferences in future outputs. This adaptive learning happens during use, complementing its initial offline training. Additionally, the developers emphasize ethical safeguards and transparency, meaning the system is designed to adjust its actions to avoid unsafe outcomes and to align with human intentions as it gains experience.
\end{itemize}

\begin{figure}[h]
    \centering
\includegraphics[width=1\textwidth]{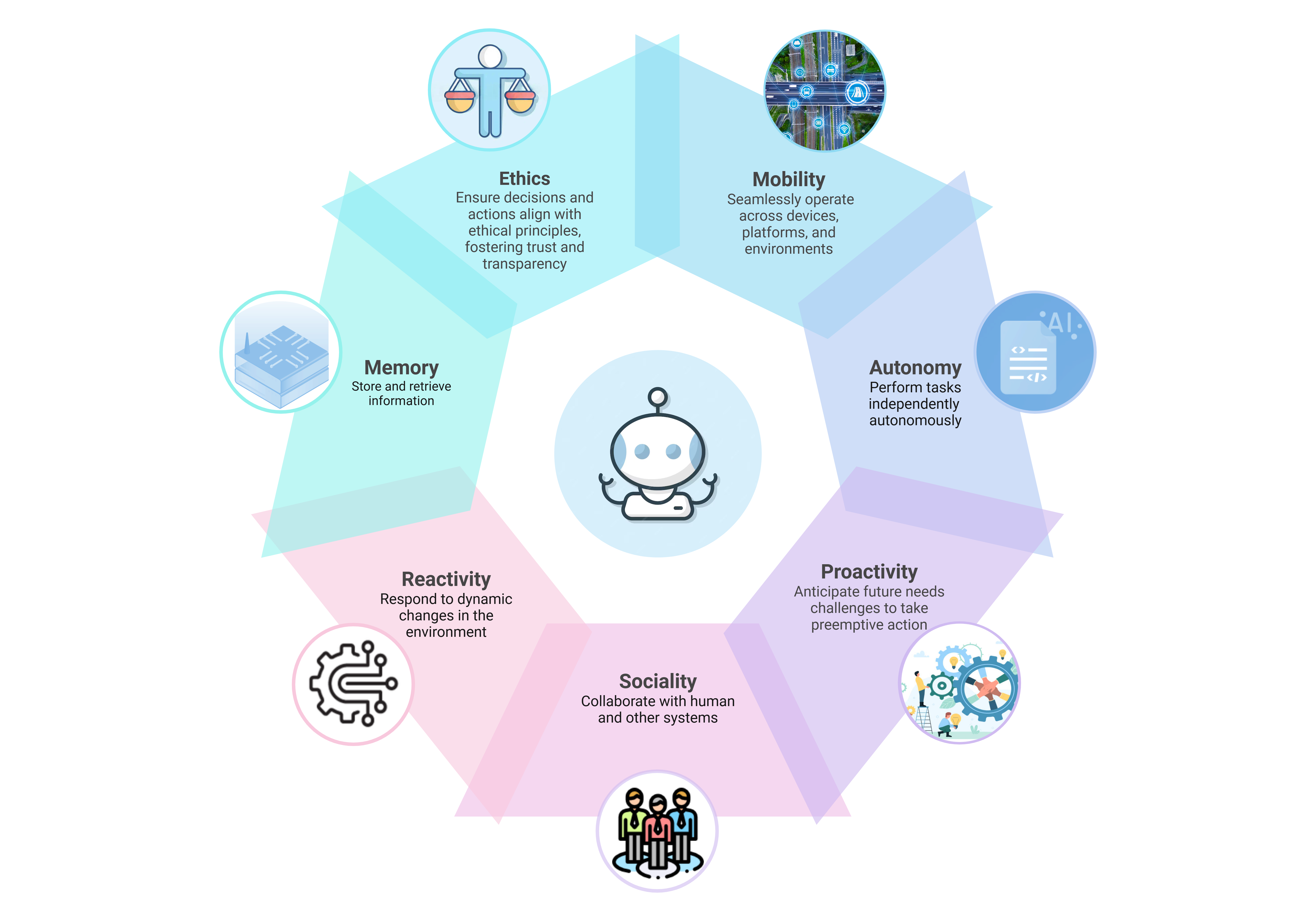}
\caption{Unique Features and Capabilities}
\end{figure}

In summary, Manus AI's inner workings combine a powerful general AI model with a clever agent framework that enables autonomous operation. Through specialized sub-agents for planning and verification, reinforcement learning for decision-making, multi-modal and tool-using proficiencies, and adaptive behavior, Manus achieves a level of autonomy and versatility that is at the cutting edge of AI technology. These technical foundations empower the wide-ranging applications of Manus AI discussed in the next section.

\section{Applications in Various Industries}
One of the most compelling aspects of Manus AI is its potential to transform numerous industries by automating and augmenting complex tasks. Because it is not confined to a single domain, Manus can be deployed wherever there is a need for intelligent decision-making and task execution. Below we explore how Manus AI can be applied in a variety of sectors, highlighting use cases in healthcare, finance, robotics, entertainment, customer service, manufacturing, education, and more. In each of these, Manus's combination of data analysis, reasoning, and autonomous action has the capacity to improve efficiency and unlock new capabilities.

\begin{figure}[h]
    \centering
\includegraphics[width=1\textwidth]{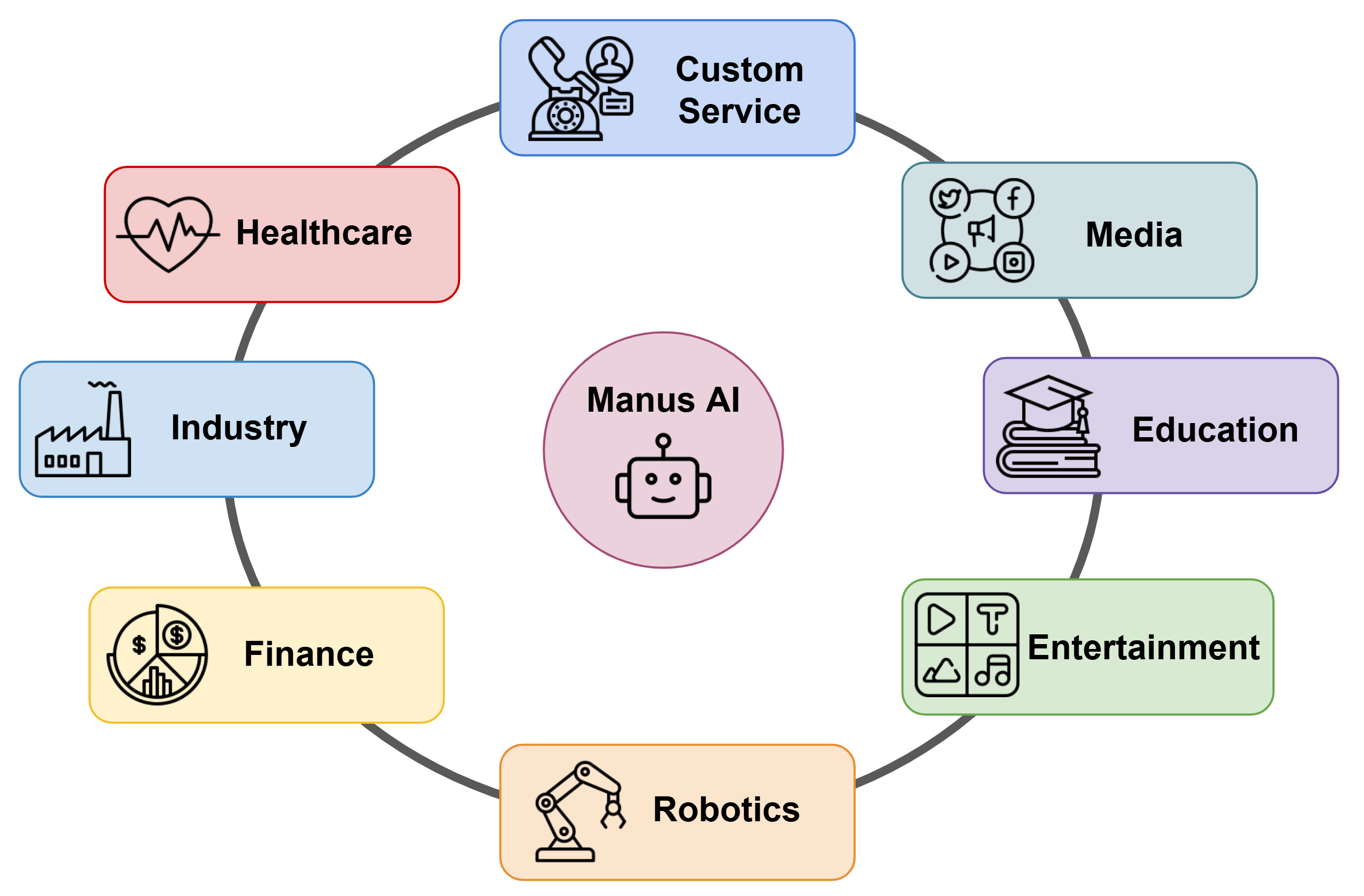}
\caption{Applications in Various Industries}
\end{figure}

\subsection{Healthcare}
In healthcare, Manus AI could serve as a powerful assistant to medical professionals and researchers. Its multi-modal abilities enable it to analyze patient records, medical literature, and even diagnostic images in tandem. For example, Manus could review a patient's history, lab results, and radiology scans to assist doctors in diagnosing complex conditions, providing a second opinion with supporting evidence from relevant medical data. Manus's long-term memory and analytical skills can potentially improve diagnostic accuracy by cross-referencing comprehensive patient information; by continuously learning from new cases, it might reduce oversight errors in interpreting results. 

Beyond diagnostics, Manus AI can contribute to \textbf{personalized treatment planning}. It can synthesize information from vast databases of medical knowledge and patient-specific factors (such as genomics or lifestyle) to propose tailored treatment options. For instance, given a cancer patient’s profile, Manus could collate the latest research on effective treatments for that cancer subtype, cross-reference clinical trial results, and provide oncologists with ranked recommendations for therapy, all annotated with source evidence. This aligns with the vision of precision medicine, where AI helps identify the right treatment for the right patient by considering many variables simultaneously. 

Another promising application is in drug discovery and biomedical research. Manus AI's autonomous research capabilities mean it could formulate and test hypotheses by mining scientific papers and databases. A pharmaceutical company could task Manus with finding novel drug targets for a disease: Manus would scan millions of publications, identify patterns in biological pathways, propose potential targets, and even design virtual screening experiments~\cite{lu2022cot}. Its ability to reason across modalities (textual hypotheses, chemical structures, experimental data) and plan experiments could dramatically accelerate the R\&D process in medicine.
Another promising application is in drug discovery and biomedical research. Manus AI's autonomous research capabilities mean it could formulate and test hypotheses by mining scientific papers and databases. A pharmaceutical company could task Manus with finding novel drug targets for a disease: Manus would scan millions of publications, identify patterns in biological pathways, propose potential targets, and even design virtual screening experiments. Its ability to reason across modalities (textual hypotheses, chemical structures, experimental data) and plan experiments could dramatically accelerate the R\&D process in medicine~\cite{newton2022applying, ji2024rag, ji2024mitigating, le2025medllm}.

Finally, Manus can play a role in \textbf{clinical operations} and patient care. As an AI assistant, it could handle routine but time-consuming tasks like writing medical reports or summarizing doctor-patient conversations, allowing clinicians to focus more on direct patient interaction. It might operate as a 24/7 virtual health agent that answers patient questions, monitors symptoms via connected devices, and alerts human providers when intervention is needed. Such an AI agent, capable of autonomous monitoring and decision support, could improve healthcare delivery by augmenting an overburdened workforce~\cite{alowais2023revolutionizing}.

\subsection{Finance}
The finance industry, with its huge volumes of data and critical need for fast, accurate decisions, is ripe for disruption by general AI agents like Manus. One key application is in \textbf{algorithmic trading and investment analysis}~\cite{10.1145/2500117, ni2024earnings}. Manus AI can continuously ingest financial news, market data, and historical trends, using that information to autonomously formulate trading strategies or investment recommendations. Unlike conventional trading algorithms that follow fixed rules, Manus can dynamically adjust strategies as new information arrives—for example, it might detect a subtle change in consumer sentiment from social media and decide to re-balance a portfolio before competitors do. In a demonstration of its financial acumen, Manus has been shown to analyze stock data, generate charts of key indicators, and produce professional-grade analyst reports complete with actionable insights~\cite{PandaYoo2025}. Such comprehensive analysis would normally require a team of human analysts; Manus can do it in a fraction of the time and update its findings in real time as conditions change.

In the realm of \textbf{risk management and fraud detection}, Manus AI offers significant advantages. Financial institutions struggle with detecting fraudulent transactions or assessing credit risks quickly enough~\cite{imi2025aifraud}. Manus can be tasked with monitoring thousands of transactions per second, identifying anomalous patterns that suggest fraud, and autonomously initiating protective measures (like blocking a transaction or flagging an account) much faster than manual review. Its adaptive learning means it can evolve with emerging fraud tactics. Similarly, for credit and risk assessments, Manus could integrate diverse data (customer financial history, macroeconomic indicators, even news about that customer's industry) to make granular risk predictions, improving on traditional credit scoring models. Because Manus can explain the factors behind its decisions, it can help risk officers understand the rationale for a flagged risk, satisfying regulatory demands for transparency.

Another financial application is in \textbf{customer service and personalized finance}. Manus AI could serve as a financial advisor chatbot that not only chats with customers but actually takes actions on their behalf. For instance, a customer might ask, ``Help me optimize my monthly budget and invest the surplus." Manus could analyze the person's spending patterns (by accessing transaction data with permission), identify areas to save, and automatically move funds into an investment account, selecting appropriate investments based on the customer's profile and goals. All of this could be done autonomously while keeping the customer informed, effectively acting as a personal financial planner that works continuously in the background.

\subsection{Robotics and Autonomous Systems}
While Manus AI exists primarily as a software agent, its capabilities can extend into the physical realm when paired with robotic systems. In robotics, Manus can function as the high-level “brain” that gives intelligent direction to machines. One application is in \textbf{industrial automation}, where Manus oversees fleets of robots on a factory floor. Because it can plan and coordinate complex sequences of actions, Manus could dynamically assign tasks to different robots, schedule their activities to optimize throughput, and adapt plans on the fly if one robot encounters a problem. For example, if a manufacturing robot goes down for maintenance, Manus would detect the issue and immediately reroute tasks to other machines or adjust the assembly sequence to prevent an assembly line halt. Its ability to integrate real-time sensor data means Manus can make context-aware decisions to keep operations running smoothly.

Another domain is \textbf{autonomous vehicles and drones}~\cite{bathla2022autonomous, ding2024confidence, wang2024data, wang2024fine}. Manus AI’s decision-making algorithms, especially its reinforcement learning backbone, are well-suited for navigation and control problems. In principle, Manus could serve as the central AI for a self-driving car network, processing traffic data, mapping information, and even verbal passenger requests to plan safe and efficient driving routes. It would execute control commands (through the car’s interface) and verify outcomes, analogous to how its Execution and Verification agents work in digital tasks. The human-like reasoning component helps in scenarios that need judgment—such as negotiating an unfamiliar construction zone or deciding how to adjust when an emergency vehicle approaches. Similarly, a fleet of delivery drones could be managed by Manus AI, which would optimize their routes, handle exceptions (like a drone encountering bad weather) by recalculating missions, and learn from each delivery to improve performance over time.

Crucially, Manus can also facilitate \textbf{human-robot collaboration}~\cite{semeraro2023human}. Many robots lack sophisticated on-board intelligence and rely on either pre-programmed routines or manual control for complex tasks. By giving such robots access to Manus AI, they gain a form of common sense and high-level understanding. Consider a scenario in a hospital: a service robot is tasked with fetching items for nurses. With Manus, the robot can understand a request like ``We need more IV stands in Room 12, and then take this medication to Room 7 if the patient is awake." Manus would break this down: navigate to storage for IV stands, prioritize if multiple tasks conflict, interpret patient status from hospital databases to know if the patient in Room 7 is ready for medication, and so forth. It essentially allows robots to follow multi-step spoken or written instructions and carry them out intelligently, asking for clarification only when necessary.

Early experiments integrating large language models with robotics support this vision. Researchers have shown that language models can translate high-level instructions into low-level robotic actions, aiding human-robot task planning~\cite{rana2023sayplan}. With a system like Manus overseeing robots, we move closer to general-purpose home or workplace robots that can be given abstract goals (“clean up this room and then set the table for dinner”) and execute them reliably by combining vision, manipulation, and reasoning. This could revolutionize sectors from warehouse logistics to eldercare, where flexible automation is in high demand.

\subsection{Entertainment and Media Production}
The entertainment industry stands to be profoundly influenced by AI agents like Manus, which can contribute to creative processes and production workflows. In \textbf{game development}, Manus AI could be used to design more intelligent and adaptive non-player characters (NPCs) or even entire game narratives~\cite{nvidia2024ace}. Game designers could specify world settings and objectives, and Manus would autonomously generate quest lines, dialogues, and dynamic events, effectively co-creating game content. Because Manus can simulate decision-making, NPCs powered by Manus could exhibit human-like strategic behavior or dialogue that evolves based on player actions, leading to games with unprecedented depth and replayability.

In \textbf{film and content creation}, generative AI is already emerging as a tool for script writing, visual effects, and editing~\cite{mims2025russo, deng2024composerx, ding2024enhance}. Manus AI takes this further by acting as a coordinator and creator in the production pipeline. For instance, a film writer could ask Manus to draft several plot outlines given a premise; Manus would not only write summaries but might also suggest key scenes and even camera angles, integrating knowledge of what makes a compelling story. In post-production, an AI like Manus could autonomously perform tasks such as editing raw footage into a coherent sequence according to a desired pacing, or generating placeholder special effects and then refining them based on director feedback. Manus's multi-modal generation means it could create storyboards (as images) from a text script, or propose music for a scene after analyzing its emotional tone.

Another area is \textbf{personalized entertainment}. Because Manus can understand individual preferences, it could curate media or even generate custom content on the fly. Imagine an interactive storytelling app~\cite{riedl2013interactive} where Manus is the storyteller: it takes a user's inputs (preferred genre, characters they like) and spins up a personalized short story or even a short animated movie by controlling generative models for images and voices. As the user reacts or provides feedback, Manus adjusts the narrative, essentially improvising a film or game tailored to one person. This kind of AI-directed experience blurs the line between creator and audience, opening up new entertainment formats.

Moreover, in media production environments, Manus can help with supporting tasks that are often time-consuming: subtitling and translating content, generating marketing materials (trailers, posters) from source content, analyzing viewer feedback and box office data to inform sequels or edits. An agent that autonomously sifts through audience comments or critiques and then suggests concrete improvements for a show would be extremely valuable. Some studios are already using AI to provide data-driven predictions on how unusual story elements will land with viewers~\cite{Davenport2023}. An AI like Manus could take those predictions and directly implement changes in the script or edit, creating a more efficient feedback loop.

While creative fields have understandable reservations about AI, Manus AI’s role in entertainment can be seen as a powerful assistant—speeding up mundane tasks and offering a wellspring of ideas—while leaving final creative judgments to human artists. The net effect could be faster production timelines and new forms of interactive content that were previously impractical to produce.

\subsection{Customer Service and Support}
Customer service is an industry that has rapidly adopted AI in the form of chatbots and virtual assistants, and Manus AI represents the next leap for this domain. Traditional customer service bots can answer FAQs or do simple ticket routing, but Manus can handle far more complex interactions and even execute service tasks start-to-finish. As a \textbf{chatbot}, Manus would be highly conversational and context-aware, remembering earlier parts of a dialogue and handling multi-turn inquiries with ease. But it would also be able to take actions on behalf of the customer: for example, a customer might contact support saying their smart home device isn't working. Manus could walk through troubleshooting steps conversationally and simultaneously interface with diagnostic tools in the background (checking the device's status online, pushing a firmware update, etc.). If a return or repair is needed, Manus could autonomously initiate that process—filling out a return authorization, scheduling a pickup, and confirming with the customer—all within the same chat session.

The benefit of such autonomy in customer service is significantly improved resolution time and consistency. Studies have shown AI-driven support can lead to faster resolution and round-the-clock availability, with one analysis reporting a 3.5x increase in support capacity for businesses using AI solutions. Manus AI can not only offer 24/7 service, but handle many issues without ever needing a human agent, freeing human representatives to focus on the most challenging cases that truly require empathy or complex judgment. Because Manus can integrate with internal company databases and knowledge bases, it can retrieve a customer's purchase history, account status, and relevant policies instantly, allowing it to personalize interactions and solve issues more efficiently than a human who must lookup information.

In addition to reactive support, Manus enables \textbf{proactive customer service}. For instance, it might monitor user account activity or device logs (with permission) to predict issues. If Manus detects that a user is frequently encountering an error in a software product, it could reach out to offer help or silently implement a fix. In e-commerce, Manus could act as a personal shopping assistant that not only recommends products but handles the entire purchasing process via conversation (“I found a better price for this item at another store and placed the order for you, shall I proceed?”).

There is also an application in \textbf{training and assisting human agents}. Manus can observe interactions between customers and human support staff (with appropriate privacy safeguards) and provide real-time suggestions to the human agent on how to resolve issues or upsell services, based on what it has learned from past interactions. It can also be used to train new support staff by simulating customer queries of varying difficulty and providing feedback.

One challenge in customer service is maintaining a high level of quality and empathy, which purely automated systems can struggle with. Manus’s advanced language model and context retention help it to handle nuanced queries with appropriate tone. However, companies would likely use Manus in a hybrid approach: the AI handles routine queries fully and assists with complex ones, with an easy escalation path to humans when needed. This approach yields the best of both worlds—speed and efficiency from the AI, and human touch where it matters. As AI continues to improve, a system like Manus could eventually resolve the majority of customer issues instantly, fundamentally changing how customer service centers operate.

\subsection{Manufacturing and Industry 4.0}
Manufacturing is undergoing a digital transformation often referred to as Industry 4.0, and AI agents such as Manus can be at the heart of this evolution. One key application is \textbf{predictive maintenance}~\cite{zonta2020predictive, jin2025adaptive, yang2024hades, yang2025research, xu2024towards, xu2024restful}. Factory equipment and machines generate a wealth of sensor data that, if analyzed properly, can predict when a part is likely to fail or when maintenance is needed. Manus AI can autonomously monitor this data in real time and detect subtle signals of wear and tear—perhaps a vibration pattern in a motor or a slight temperature increase in a turbine bearing. By catching these early, Manus can schedule maintenance before a breakdown occurs, thus avoiding costly downtime. According to a PwC study, manufacturers using AI-based predictive maintenance have seen up to a 9\% increase in equipment uptime and 12\% reduction in maintenance costs~\cite{PwC2018}. Manus's ability to both analyze data and act (by generating work orders or alerts to technicians) makes it a full-cycle solution for maintenance optimization.

In \textbf{process optimization}, Manus can serve as a real-time decision agent on the production line. Modern manufacturing involves complex coordination of supply chains, production schedules, and quality control~\cite{yadav2024industrial}. Manus could take in live data about raw material availability, machine performance, and order deadlines, and then dynamically adjust the production plan. For example, if a supply shipment is delayed, Manus might re-order the assembly sequence to prioritize products that do have all components ready, or instruct machines to switch to a different batch that can be completed, thereby keeping the factory productive. Similarly, Manus can monitor quality metrics (via sensors or machine vision on the line) and if it detects the production of substandard units, it can adjust machine settings or call for human inspection. Over time, by learning from output data and yields, Manus could continuously refine how machines are configured, pushing production efficiency to new highs that would be hard to achieve with static, pre-programmed logic.

Another significant area is \textbf{supply chain and logistics management}. A manufacturing AI agent could seamlessly connect to suppliers, track inventory levels, and even negotiate orders or delivery schedules. Manus might predict that a certain component will run out in two weeks based on the current burn rate and automatically place an order while also arranging the most cost-effective shipping. In warehousing, Manus can guide autonomous forklifts or robots to manage inventory placement and order fulfillment optimally, as discussed in the robotics section. By having a global view of the entire manufacturing ecosystem and the autonomy to make decisions, Manus AI can eliminate much of the latency and inefficiency in supply chain responses. Manufacturers using such AI could react to market changes or disruptions almost instantly—for instance, scaling back production ahead of a forecasted dip in demand, or quickly sourcing alternatives if a supplier fails—thus saving money and staying agile.

One can envision a future “lights-out” factory where human oversight is minimal: Manus AI schedules production, runs the robots, ensures maintenance, manages supply chain logistics, and only pings humans when a strategic decision or a truly novel situation arises. While completely autonomous factories are still rare, the components of this vision are falling into place, and Manus exemplifies the kind of general AI agent that could coordinate all these pieces under one umbrella of intelligence.

\subsection{Education}
Education is another field where Manus AI's capabilities can be transformative by enabling highly personalized and interactive learning experiences. As a \textbf{tutor or teaching assistant}, Manus can adapt to the learning style and pace of each student. It can explain difficult concepts in multiple ways, generate practice problems tailored to a student's weak spots, and provide instant feedback on answers. Unlike a human teacher who must divide attention among many students, Manus could potentially give one-on-one tutoring to every student simultaneously. It can remember each student’s progress in detail, ensuring that no concept is left misunderstood. For example, if a student is struggling with a calculus problem, Manus can recognize confusion from the student's queries or mistakes and switch strategies—perhaps using a visual demonstration or drawing on an analogy from a subject the student excels in—to make the concept click.

This goes hand-in-hand with \textbf{personalized curriculum generation}~\cite{duan2019automatic}. Manus AI can design a learning plan optimized for an individual's goals and current knowledge. Suppose a student wants to learn programming for web development. Manus can assess the student's current math and logic skills and then create a sequence of lessons and projects that teach the necessary programming concepts, adjusting difficulty as the student improves. It can integrate multimedia (text, code examples, video explanations) and even interactive coding environments as part of the curriculum. As the student advances, Manus continuously updates the learning plan, maybe introducing more challenges or circling back to reinforce earlier topics that were troublesome.

For teachers and educational content creators, Manus can serve as a \textbf{content generation and grading assistant}~\cite{lee2024college}. It can generate quiz questions or exam papers covering specific topics with varying difficulty levels. It can also grade free-form answers or essays by applying rubrics—providing not just a score but also detailed feedback. This is particularly useful in large open online courses or education at scale, where subjective grading is a bottleneck. Additionally, Manus could help in creating illustrative examples, diagrams, or even educational games on the fly to help explain topics, functioning like a creative partner for educators.

The classroom of the future might involve each student having an AI tutor like Manus on their device or available in the classroom. The AI tutor can handle routine instruction and practice, while the human teacher focuses on higher-level mentoring, motivation, and social-emotional learning. AI like Manus can also assist students with disabilities by offering tailored support—for instance, converting lesson content to more accessible formats or giving extra practice in areas of difficulty—thus supporting inclusive education.

It is worth noting that early forms of AI tutors have shown promise in improving learning outcomes by providing students with immediate, individualized feedback. Manus’s advanced reasoning and memory could amplify these benefits, as it not only answers questions but can figure out why a student made a mistake and address the root cause. As a concept demonstration, an AI agent like Manus might generate personalized learning plans for students and provide on-demand explanations, effectively acting as a tireless teaching aide. The potential scale of impact in education is huge: Manus-like AI assistants could democratize access to high-quality tutoring and help reduce educational inequities by giving every student a personal tutor attuned to their needs.

\subsection{Other Fields}
Beyond the industries detailed above, Manus AI’s general capabilities open opportunities in many other areas:
\begin{itemize}[leftmargin=*]
    \item \textbf{Legal Services}: Manus can function as a paralegal aide by reviewing lengthy legal documents and contracts, highlighting key points or inconsistencies, and even drafting initial versions of legal briefs. Given a query, it can research case law and compile relevant precedents. This automation can drastically reduce the time lawyers spend on research and document preparation. Demonstrations have shown Manus handling legal contract review from end-to-end, ensuring no clause is overlooked~\cite{AITech2025}.
    \item \textbf{Human Resources}: In recruitment, Manus AI can screen résumés and job applications at high speed, identifying the most suitable candidates based on a company's criteria. It doesn't just keyword-match; Manus can interpret descriptions of experience and skills contextually, making judgments much like a human recruiter. One use case had Manus parse and evaluate a stack of résumés, extracting key qualifications and ranking applicants efficiently~\cite{PandaYoo2025, xiang2024neural}. Additionally, Manus can assist in employee training by providing personalized learning modules and answering policy-related questions for staff.
    \item \textbf{Real Estate and Planning}: Manus can automate real estate analysis by scanning property listings, comparing them against a buyer's preferences and budget, and producing a shortlist of best matches complete with pros/cons and investment outlooks~\cite{yang2024using}. It can also generate property valuation reports and even draft offer letters or rental agreements. As noted in one example, Manus was tasked with real estate research and managed to compile detailed reports on available properties meeting specific criteria, saving clients from hours of search and comparison~\cite{PandaYoo2025}.
    \item \textbf{Scientific Research}: Researchers can use Manus as an analytical assistant to simulate experiments or analyze experimental data. For instance, in a physics lab, Manus could control equipment via software, gather data, fit it to theoretical models, and suggest interpretations. It can also automatically write up initial drafts of research papers by organizing the experimental context, method, results, and related work from references it has read. Such capabilities could accelerate the research cycle in fields from biology to engineering~\cite{ wang2024twin}.
    \item \textbf{Public Sector and Smart Cities}: Governments and city planners might use Manus AI to optimize public services~\cite{kalyuzhnaya2025llm}. For example, Manus could analyze traffic patterns, public transit usage, and events schedules to optimize traffic light timings or recommend changes in transit routes in real time, improving urban mobility. In public health, Manus could monitor epidemiological data and coordinate responses to health crises by suggesting where to allocate resources. Its autonomy means it could continuously manage and adjust city systems (water, power distribution, emergency services deployment) based on current data, aiming for maximal efficiency and rapid response to incidents.
\end{itemize}

These examples only scratch the surface. Virtually any field that involves complex decision processes, large datasets, or multi-step workflows could leverage Manus AI to some extent. The common thread is that Manus brings a combination of cognitive skills (understanding context, learning, reasoning) and the ability to act (through tool usage or executing instructions). This makes it a kind of \emph{universal problem-solver assistant} that can be pointed at tasks in any domain and, with minimal adaptation, start contributing productively.

\section{Comparison with Other AI Technologies}
Manus AI's emergence comes at a time when many organizations are racing to build more advanced AI systems. It stands out in comparison to existing technologies from leading AI labs like OpenAI, Google DeepMind, and Anthropic, among others. In this section, we analyze how Manus differs from and potentially surpasses these contemporaries, highlighting unique aspects as well as any trade-offs.

\subsection*{Manus AI vs. OpenAI's GPT-4 and Agents}
OpenAI's GPT-4, released in 2023, is one of the most well-known AI models, demonstrating remarkable abilities in language understanding and generation~\cite{achiam2023gpt}. GPT-4 can solve problems, write code, and hold conversations at a high level of fluency. However, GPT-4 (and its publicly deployed form, ChatGPT) operates primarily as an interactive assistant that replies to user inputs. It does not inherently have the capacity to execute multi-step plans autonomously without continuous prompting. Manus AI was built to overcome this limitation. Unlike GPT-4 which provides suggestions or information, Manus is designed to take initiative and carry out tasks end-to-end~\cite{LLMHacker2025}. For instance, GPT-4 might tell you how to analyze a dataset, but Manus will actually perform the analysis, create charts, and deliver a report without further prompting.

In internal evaluations like the GAIA benchmark~\cite{mialon2023gaia}, Manus AI demonstrated stronger performance on practical task execution than GPT-4~\cite{LLMHacker2025}. GPT-4, augmented with plug-in tools, has started to move in Manus's direction by allowing limited web browsing or code execution, but those features are not as seamlessly integrated or generally capable as Manus's tool use. Manus effectively has the tool-using and action-taking parts woven into its core architecture rather than tacked on. This means Manus plans when and how to use tools as part of its natural reasoning process, whereas GPT-4 relies on external orchestration to do something similar. Indeed, Manus achieved higher task completion rates on GAIA than a version of GPT-4 with plug-ins enabled, which scored significantly lower~\cite{LLMHacker2025}.

Another distinction is accessibility and openness. OpenAI's models, while proprietary, are widely available via APIs or consumer-facing apps, enabling extensive independent evaluation by the community. Manus AI, in contrast, has been kept relatively closed (invitation-only beta at this stage). This means independent benchmarks are limited to what the developers report. Some experts have expressed skepticism about Manus's claimed superiority until more public testing is possible. Nonetheless, the available evidence (demos and benchmark reports) indicates Manus's novel architecture gives it an edge in autonomy that even GPT-4 doesn’t have out-of-the-box. 

It's also worth noting that OpenAI has been developing its own agent-like frameworks (such as the open-source \textit{AutoGPT}~\cite{AutoGPT} or internal projects to make GPT models more agentive). Manus can be seen as part of the same paradigm shift, but it appears to have leapfrogged into a more advanced implementation first. If GPT-4 is an exceptional problem-solver when guided, Manus is an independent problem-solver that can figure out what needs doing with minimal guidance~\cite{du2024embracing}.

\subsection*{Manus AI vs. Google DeepMind's AI}
Google's DeepMind division has produced some of the most impressive AI breakthroughs, from AlphaGo (which mastered the game of Go)~\cite{silver2016mastering, silver2017mastering} to AlphaFold (which solved protein folding)~\cite{jumper2021highly,senior2020improved}, and they have experimented with generalist models like \textit{Gato} that can perform multiple kinds of tasks. DeepMind is also collaborating with Google Brain on next-generation models (e.g., the upcoming multimodal model \textit{Gemini}). However, many of DeepMind's systems, until now, have been highly specialized or confined to specific environments (like games or simulations) rather than being user-facing general agents.

Where Manus AI distinguishes itself is in being a broad, user-interactive agent capable of open-ended tasks in the real world. DeepMind's \textit{Sparrow}~\cite{deepmind2024saferdialogue} and other chatbots focus on dialogue and factual accuracy, but they do not execute physical or digital tasks for the user. A more analogous DeepMind project might be their research on adaptive agents that can use tools (DeepMind has published work on combining language models with tool use and reasoning as well). However, those are research prototypes, whereas Manus is positioned as a deployable product.

DeepMind has a track record of emphasizing fundamental research and optimal performance (for example, AlphaGo was extremely optimized for Go). Manus, by comparison, might not match a specialized DeepMind model in a narrow domain (for instance, it won't play Go as well as AlphaGo), but it brings a \emph{breadth} of competence that DeepMind's individual models don't have. It is akin to the difference between a champion sprinter and a decathlete; Manus is trying to be a decathlete in the AI sense.

One area to compare is reasoning and safety. DeepMind models often incorporate heavy doses of reinforcement learning and have excelled at planning in simulated environments (like game strategies). Manus also uses reinforcement learning for real-world task planning~\cite{ChinaOrg2025}, effectively bringing that paradigm into more practical settings. Regarding safety, DeepMind has been cautious — for instance, Sparrow was designed with constraints to avoid unsafe answers. Manus claims to implement ethical constraints and transparency as well, but until more public data is available, it is hard to gauge how its safety mechanisms compare to DeepMind’s alignment work. It is likely that Manus's developers have integrated rule-based filters or reward signals to discourage undesirable behavior, but OpenAI and DeepMind have had the advantage of iterative refinement in the public eye.

In summary, while DeepMind (and Google’s AI efforts) might have more pure research power and resources behind them, Manus’s significance is in showing a working general AI agent tackling everyday tasks now. It stands as a proof of concept that the gap between experimental AI and practical general agents is closing. It remains to be seen if DeepMind’s upcoming systems (like Gemini) will incorporate similar agentive features and how they will stack up against Manus.

\subsection*{Manus AI vs. Anthropic's Claude and Others}
Anthropic, an AI safety and research company, has developed the \textit{Claude} series of language models, which are direct competitors to OpenAI's GPT models. Claude is known for its large context window and a training focus on helpfulness and harmlessness through a method called Constitutional AI~\cite{bai2022constitutional}. When comparing Manus AI to Anthropic’s Claude, one notes a similar dichotomy as with GPT-4: Claude is an extremely capable conversational model, but it does not natively perform multi-step tool-using tasks without external frameworks. Manus has been touted as surpassing Anthropic's Claude on combined benchmarks of reasoning and action (being described as having capabilities beyond “Claude + tool use” in some commentaries). This is plausible given Claude was not primarily designed as an autonomous agent.

Another perspective is that Manus was described as a fusion of ``OpenAI's DeepResearch~\cite{openai2024deepresearch} and Claude's computer-use capabilities~\cite{anthropic2024-computer-use}," implying it took inspiration from strengths of both OpenAI and Anthropic models. Enthusiasts suggested that Manus combined OpenAI-level reasoning with Claude-like tool use, plus the added ability to write and execute its own code — resulting in what one observer called a “monster” of AI capability that arrived sooner than expected.

Outside of Anthropic, there are other emerging AI systems. For example, new startups and big tech companies are launching their own general AI agents: Amazon's experimental \textit{Nova} project~\cite{amazon-nova}, or Elon Musk's \textit{xAI} initiative with a model called Grok, are aimed at similar goals. Manus’s advantage of being first to showcase a fully autonomous general agent could be challenged as these players catch up. That said, according to industry commentary, compared to competitors like xAI's Grok and Anthropic's Claude, Manus's autonomy and task completion capabilities are seen as differentiating advantages in this early stage~\cite{Aibase2025}. Manus has set a high bar that others will now aim for.

It's also worth mentioning smaller but notable contributors: H2O.ai’s h2oGPT-based agent~\cite{h2oai_h2ogpte} was leading the GAIA benchmark before Manus, demonstrating that even less prominent players can innovate. Manus overtook that score, highlighting the rapid progress in this area. In China, another project called \textit{DeepSeek} gained attention earlier for an AI chatbot that became very popular~\cite{liu2024deepseek-v3}. Manus is often compared as the next ``DeepSeek moment," but focusing on autonomy rather than just conversation. The Chinese tech ecosystem, backed by strong investment, means Manus might soon face domestic competition as well.

In summary, the competitive landscape is vibrant. Manus AI sets itself apart with a focus on true autonomy and generality, whereas most other AI products currently excel either in conversational intelligence (like GPT-4, Claude) or in narrow domain mastery (like AlphaGo). Manus attempts to do both—to understand and to act—which is why it is seen as a step toward general AI agents. It is not necessarily that Manus has a fundamentally different kind of AI “brain” — it still relies on large language model technology similar to others — but it has an innovative system design that makes that brain much more usefully applied. If Manus's approach proves effective, we can expect other AI leaders to integrate more agent-like behaviors into their systems. Manus has, in a sense, thrown down a gauntlet: showing what a focused team can accomplish by tightly integrating existing AI techniques (LLMs, RL, tool interfaces) into a single agent. The ultimate winners are likely to be users and businesses, who will gain access to increasingly powerful AI agents from multiple sources.

\section{Pros and Cons of Manus AI}
As an advanced AI agent, Manus AI exhibits a number of significant strengths, while also presenting certain limitations and challenges. Understanding these pros and cons is crucial for evaluating Manus's overall impact and guiding future improvements.

\subsection*{Strengths and Advantages}
\textbf{Autonomy and Efficiency}: The foremost strength of Manus AI is its ability to operate autonomously once given a goal. This can dramatically increase efficiency in completing tasks. Users do not need to micromanage or break tasks into sub-tasks—Manus handles the entire process. In practical terms, this can save time and labor; tasks that might take a team of humans hours or days of coordination could be done by Manus in minutes or seconds. For example, generating a comprehensive market research report might normally involve researchers gathering data, analysts interpreting it, and writers compiling the document. Manus can perform all these stages by itself, from web scraping data to analysis to writing up results, thus collapsing workflows.

\textbf{Versatility}: Manus’s generalist design and multi-modal competence make it highly versatile. It can transition from one domain to another without needing to be re-engineered. This “jack of all trades” ability means a single instance of Manus AI could assist multiple departments of a company in different ways, or a single user in various aspects of life. Versatility also future-proofs Manus to an extent—if new tasks or tools emerge, Manus’s architecture is built to incorporate them (through additional training or integration) relatively easily, rather than having to create a new model from scratch.

\textbf{State-of-the-Art Performance}: Manus has demonstrated state-of-the-art performance on challenging benchmarks, as discussed earlier (GAIA results surpassing other models). While benchmarks aren’t everything, they indicate that Manus’s reasoning and problem-solving abilities are at the cutting edge. Its creators report that it achieves top-tier results even on the hardest task categories, outperforming contemporary AI models~\cite{AITech2025,MalayMail2025}. In user-facing trials, many have been impressed by Manus’s ability to handle tasks that other AI systems struggle with (like deeply multi-step queries or combining knowledge from disparate sources). Being ahead of competitors technologically gives Manus a first-mover advantage in the market for autonomous AI agents.

\textbf{Tool Use and Integration}: Manus’s adeptness at integrating with external systems is a huge practical advantage. It can plug into existing software ecosystems, meaning it can be deployed to work with a company's current applications rather than requiring a whole new platform. Businesses can, for instance, connect Manus to their databases, CRM systems, or DevOps pipeline and have it execute actions. This integrated approach turns Manus into an “AI employee” of sorts that can actually press the buttons and not just advise. Competing AI that lack this integration act more like consultants that tell you what to do, whereas Manus can be the hands that do the work.

\textbf{Continuous Improvement}: Manus AI is designed to learn from interactions. Over time and with more usage, it can become even more personalized and fine-tuned to its environment. This means Manus deployments have the potential to improve without major updates, as the system adapts to the specific data and preferences it encounters. Such continual learning is powerful; it’s akin to an employee gaining experience on the job. Of course, this requires careful handling to avoid drifting from correctness, but in controlled ways it means Manus today could be better than Manus yesterday if it's learning from its mistakes. Moreover, the developers of Manus will likely refine the model with broader data and user feedback, addressing weaknesses and expanding knowledge, so the core AI will keep getting smarter and more capable.

\textbf{Global Reach and Language Support}: Given its training on large-scale data, Manus AI likely supports multiple languages and can serve globally. This broad language capability means Manus can be beneficial in diverse linguistic contexts, an advantage in international applications compared to tools that might be English-centric. It can potentially mediate multilingual communication (e.g., translating while analyzing content) which adds to its utility in globally operating organizations.

\subsection*{Limitations and Challenges}
\textbf{Lack of Transparency}: One challenge with Manus AI, as with many deep learning-based systems, is that its decision-making process can be opaque. While it has a Verification agent that checks results, understanding exactly how Manus arrived at a complex decision can be non-trivial. This “black box” nature might concern users in high-stakes domains like healthcare or law, where being able to justify a decision is essential. The developers have stated the importance of transparency and ethical boundaries in Manus's design, but it is not clear to what extent Manus can explain itself beyond providing the output. Improving explainability (for instance, having Manus produce a rationale or audit trail for its actions in human-readable terms) is an ongoing challenge.

\textbf{Verification and Reliability}: Although Manus has an internal verifier, no AI system is infallible. There may be cases where Manus executes a plan that turns out to be suboptimal or even wrong. If the Verification agent fails to catch an error or if the data sources Manus uses are flawed, it could produce incorrect results confidently. For example, if Manus is gathering information from the web and it encounters misinformation, it might incorporate that into its analysis. Current AI models are known to sometimes “hallucinate” facts or logic. Manus’s added structure might reduce that, but not eliminate it. Therefore, handing over critical tasks entirely to Manus carries risk until it has an extensive track record. Human oversight or review may still be needed for important outputs, which partially offsets the autonomy advantage.

\textbf{Data Privacy and Security}: For Manus to function effectively, it often needs access to sensitive data (medical records, financial information, internal business documents, etc.). This raises concerns about data privacy and security. Organizations might be hesitant to plug Manus in with full access to their data silos without robust assurances that it won't misuse or leak that information. Any vulnerability in Manus’s integration (like connecting to external tools) could be a vector for cyberattacks or data breaches. Additionally, if Manus is a cloud-based service, there are the usual concerns about storing data externally. These are not unique to Manus, but its broad applicability means it will frequently face scenarios involving protected information (e.g., patient data under HIPAA~\cite{HIPAA}, consumer data under GDPR~\cite{GDPR}). Addressing these requires strong encryption, access controls, and possibly on-premise deployment options where necessary so data doesn’t leave a company's secure environment.

\textbf{Computational Resources}: Running a system as complex as Manus AI is likely computationally intensive. The multi-agent architecture and large underlying model require significant processing power, especially for real-time performance. This could translate into high operational costs or the need for specialized hardware (such as ASIC). For users, it might mean that using Manus extensively (e.g., for large-scale automation) incurs notable cloud computing expenses, which could be a barrier compared to simpler automation scripts or even human labor in some cases. Over time, as hardware improves and the model is optimized, this cost will come down, but at present, the cost and scalability of the backend might limit Manus’s deployment for extremely large-scale or latency-sensitive scenarios.

\textbf{Accessibility and Availability}: As noted, Manus AI has so far been released in a limited manner (invitation-only web preview). Currently it is not broadly accessible to all who might want to use it, which could slow the accumulation of community trust and widespread adoption. If this exclusivity continues, it may give competitors time to catch up or reduce Manus's mindshare. Additionally, if the model and agent run on centralized servers, users are dependent on the service being operational. Any downtime or outages on Manus’s platform could disrupt businesses that rely on it. In contrast, some may prefer self-hosted or offline-capable AI systems for mission-critical tasks that demand maximum uptime. Providing clear availability guarantees or offline modes is a challenge Manus's providers would need to address for enterprise acceptance.

\textbf{Ethical and Control Issues}: Granting an AI agent autonomy to execute tasks raises ethical and control considerations. Manus can act like a super-assistant, but one must be cautious about what it is allowed to do. For instance, if Manus is used in finance to execute trades and it makes a wrong judgment, who is accountable? If it's used in HR and inadvertently shows bias in hiring recommendations (perhaps reflecting biases in the training data), this could cause fairness issues. Ensuring Manus's decisions align with human values and company policies is an ongoing challenge. The developers must carefully encode constraints and monitor outputs to prevent undesirable behavior (like privacy violations, biased decisions, or unsafe actions). This is part of AI ethics. While Manus is built with an emphasis on following rules and maintaining transparency, constant vigilance is needed as the system encounters new situations. Organizations using Manus will likely need to establish guidelines for its use and have fallbacks if the AI behaves unexpectedly.

In summary, Manus AI's pros position it as a groundbreaking tool that can drive efficiency and innovation across many fields. Its cons remind us that it is not a magic infallible entity but a technology with limitations that must be managed. Overcoming issues like transparency, reliability, and security will be key to Manus AI's sustained success and acceptance. Many of these challenges are active areas of development, and we expect improvements as Manus and similar agents evolve.

\section{Future Prospects}
Manus AI represents an early leap into a new category of AI systems, and its trajectory will be shaped by both technological progress and how society chooses to embrace such agents. Looking ahead, there are several key areas where Manus AI and its successors are likely to evolve, as well as broader impacts they may have on the field of AI and on society at large.

\subsection*{Advancements in Capabilities}
In future iterations, we can expect Manus AI to expand its toolkit and refine its skills. One anticipated development is the \textbf{expansion of tool integrations}~\cite{LLMHacker2025}. Today Manus might be able to use web browsers, office applications, and coding environments; tomorrow it could seamlessly integrate with a much larger array of third-party services and hardware. For example, we might see Manus tie into engineering design software (to act as an AI CAD designer), biotech lab equipment (to function as a lab assistant controlling experiments), or personal smart home devices (acting as an AI butler for home automation). Each new integration would increase Manus's utility and domain reach.

Another area of growth is \textbf{enhanced multi-modal perception}~\cite{LLMHacker2025}. While Manus already handles text and images, future versions may achieve deeper understanding of audio (e.g., transcribing and interpreting real-time conversations or sound cues), video (e.g., analyzing live video feeds or assisting with video editing in real-time), and even haptic or spatial data (if connected to robots or IoT sensors). This would make Manus a more perceptive agent in physical environments. For instance, pairing it with security cameras could allow Manus to monitor physical premises and trigger actions (like notifying authorities or adjusting building controls) based on what it “sees.” Essentially, Manus could evolve from a mostly digital-world agent to one that also navigates and responds to the physical world.

Another likely focus is \textbf{learning and adaptation}. We might see Manus incorporate advanced online learning algorithms that let it update its knowledge base or model parameters as it encounters new data (with safety checks). If achieved, Manus could become more personalized and current without needing full retraining by its developers. Imagine a corporate Manus AI that gradually learns the specific terminology and procedures of that company over time, becoming uniquely expert in that organization's operations. Techniques like federated learning (learning from user data in a decentralized way) could be employed to maintain privacy while improving the model on the fly.

\subsection*{Wider Deployment and Use Cases}
If Manus AI continues to prove its worth, we can expect much wider deployment. In the enterprise sector, general AI agents could become as common as databases or cloud services. Companies might have an AI agent integrated into many departments handling cross-functional tasks. This could lead to \textbf{workflow redesign}: organizations may restructure around what tasks humans do versus AI agents. Routine analytical tasks might be largely handed off to AI, while humans focus on creative, strategic, or interpersonal roles. New job categories might emerge, like "AI workflow manager" or "AI ethicist," who specialize in overseeing AI agents like Manus.

For individual consumers, perhaps a future Manus-like assistant becomes a ubiquitous personal companion—far more powerful and proactive than today's voice assistants (like Siri or Alexa). It could manage one’s schedule, finances, communications, and more in an integrated way. The convenience could be profound, though it also raises questions of dependency and privacy (entrusting so much to an AI). It's quite possible that competition in this space will produce consumer-facing general agents derived from the Manus concept, each integrated into tech ecosystems from different providers.

We may also witness \textbf{collaboration between AI agents}. If many general agents exist, they might communicate to coordinate on large tasks—essentially a network of Manus instances dividing and conquering a massive problem (for example, climate data analysis or large-scale economic modeling). Standard protocols for AI-to-AI collaboration could develop. Alternatively, one Manus could consult another specialized AI as a tool, orchestrating not just software APIs but other AI services (think Manus invoking a medical diagnosis model as needed). This synergy of AI systems could amplify what each can do alone.

\subsection*{Influence on AI Research and Development}
The advent of Manus AI could significantly influence the direction of AI research. It provides a concrete demonstration that combining language models with planning, memory, and tool use yields powerful results. We will likely see more research into \textbf{agentive AI frameworks}. Competing approaches, such as those from academic labs or open-source communities, will iterate on multi-agent architectures, exploring different ways to split tasks among sub-agents or even using different cognitive architectures beyond Transformers. There may be experiments with agents that incorporate symbolic reasoning modules to improve reliability in areas like mathematics or logic.

This progress could accelerate movement toward what many consider the holy grail: \textbf{Artificial General Intelligence (AGI)}. Manus itself might not be AGI, but it points in that direction by being able to handle variety and showing a glimmer of adaptive, general problem-solving. Future research might focus on increasing the generality even more—ensuring the AI has fewer blind spots or knowledge gaps, making it better at transfer learning (applying knowledge from one domain to a completely new one), and integrating it with formal reasoning to reduce errors. Manus’s success (if it continues) will validate the concept that a system-oriented approach (multiple components + learning) can achieve more general behavior without requiring an impossibly perfect single model. This could shift some research from purely scaling models up to also composing them in smarter ways.

We might also see more emphasis on \textbf{benchmarks and standards} for AI agents. GAIA is one such benchmark; others will likely be developed to measure an AI agent’s practical usefulness, safety, and generality. Manus’s top ranking will be challenged, and competitive benchmarking will drive improvements across the industry, akin to how benchmarks like ImageNet drove rapid progress in vision models in the 2010s.

\subsection*{Societal Impact and Considerations}
The proliferation of Manus-like AI will have broad societal implications. In the workplace, as mentioned, there could be displacement of certain job functions. Tasks that are routine, data-heavy, or procedural might largely shift from humans to AIs. This doesn't necessarily mean eliminating jobs; it might transform jobs. Professionals might have an AI on their team as a junior (albeit very capable) teammate. Education and training may adapt to focus on skills that complement AI (like oversight, complex creative thinking, or emotional intelligence) rather than compete with it.

There is also the possibility of \textbf{democratizing expertise}. If everyone has access to an AI agent that is a competent lawyer, doctor, accountant, and engineer all-in-one, that could greatly reduce barriers to knowledge and services. People in remote or underserved areas could get expert advice via AI when human experts are not available. This is an optimistic outlook: AI as a great equalizer. The counterpoint is ensuring the advice is accurate and that people don't overly rely on it without proper context (e.g., misinterpreting medical guidance without a real doctor involved at some point).

From an innovation standpoint, having AI agents handle a lot of grunt work might supercharge human creativity and entrepreneurship. Imagine an individual or a small startup able to achieve what currently takes a whole company, because their AI agents handle marketing, coding, design, and logistics in the background. This could lead to a burst of innovation and productivity, as well as new business models we haven't thought of yet.

However, concerns will remain around \textbf{AI alignment and control}. As these agents become more powerful and possibly are given more autonomy (for example, managing critical infrastructure or financial systems), ensuring they remain aligned with human values is paramount. Ongoing research in AI safety will likely intensify, aiming to formally verify that agents do not act outside of allowed bounds. Manus’s developers and others might incorporate more rigorous guardrails, perhaps limiting the scope of actions in high-risk domains until confidence is extremely high. We may also see policymakers stepping in to set guidelines for autonomous AI behavior.

On the policy front, governments may start to regulate AI agents specifically. We might see certification requirements for AI used in medicine or finance, for instance. There could be discussions about whether an AI must identify itself as such when interacting (to avoid confusion or deception). Liability frameworks will need updating: if an autonomous agent causes harm, who is legally responsible? These legal and ethical frameworks will evolve as agents like Manus become integrated into daily life.

In conclusion, the future for Manus AI and similar general AI agents is one of tremendous potential coupled with significant responsibility. The next few years will likely see rapid improvements in the technology, broader adoption in many fields, and a vigorous global dialogue about how to maximize the benefits of such AI while managing the risks. Manus AI has set in motion what might be one of the most important technological shifts of the coming decade—one where AI moves from the role of a tool to that of a partner or autonomous colleague in virtually every human endeavor.

\section{Conclusion}
Manus AI stands at the forefront of a new generation of AI systems that combine understanding, reasoning, and action. In this paper, we have surveyed the landscape of Manus AI: starting from its innovative architecture that interweaves multiple specialized agents with a powerful core model, through its wide-ranging applications across industries, to its standing among contemporaries and the strengths and weaknesses that define it. Manus AI's ability to autonomously plan and execute tasks marks a significant departure from the assistive AI paradigms that have dominated in recent years. It embodies the transition toward AI that not only answers questions but delivers results.

Our exploration shows that Manus AI can potentially revolutionize fields as diverse as healthcare, finance, robotics, entertainment, customer service, manufacturing, and education. By serving as a tireless and knowledgeable assistant, it augments human capability and promises efficiency gains and innovations that are just beginning to be realized. At the same time, the comparisons with other AI leaders like OpenAI, DeepMind, and Anthropic highlight that Manus is part of a broader momentum in AI—various organizations are converging on the idea of more agentive, general AI, though with different implementations. Manus currently leads in some benchmarks of real-world problem-solving~\cite{AITech2025}, but competition will spur all players to improve, ultimately benefiting users and society.

We also delved into the pros and cons of Manus AI. Its autonomy, versatility, and performance are balanced by concerns over transparency, reliability, and the need for robust ethical guardrails. These are active areas of development. How well Manus addresses these issues will influence trust and adoption. Responsible deployment will be key to ensuring that the technology amplifies human potential without causing inadvertent harm or disruption.

Looking ahead, the evolution of Manus AI and its successors is poised to be rapid. We anticipate ongoing improvements in capability, broader deployment scenarios, and consequential impacts on work and daily life. Manus AI might be a precursor to systems that eventually qualify as a form of artificial general intelligence, albeit likely operating under human oversight and in partnership with us. Its success will inform design principles for such future AI—demonstrating the importance of features like multi-agent coordination, tool use, and continuous learning in achieving generality.

In conclusion, Manus AI can be seen as both a milestone and a harbinger. It is a milestone in that it has showcased what is possible when AI is designed to think and act in tandem, solving problems in an end-to-end fashion. It is a harbinger in that it foreshadows a near future where intelligent agents are commonplace, handling myriad tasks and collaborating with humans on complex endeavors. The arrival of Manus AI underscores the rapid progress of AI advancements and offers a glimpse into an era where the boundaries between human work and machine work become increasingly fluid.

The journey of Manus AI is just beginning, but it encapsulates many of the hopes and challenges of the AI community. If developed and deployed thoughtfully, Manus AI and systems like it have the potential to drive tremendous positive change—enhancing productivity, fostering innovation, and even helping address global challenges by providing powerful new tools for problem-solving. It also urges us to proactively address the ethical and societal dimensions of AI. The importance of Manus AI thus goes beyond its technical specifications; it invites us all to participate in shaping how such autonomous AI agents will integrate into our world. The coming years will reveal how this balance is struck, and Manus AI will undoubtedly be a central case study in that unfolding story.

\bibliographystyle{unsrt}

\bibliography{reference}
\end{document}